\documentclass[a4paper, 11pt]{article}
\usepackage[english]{babel}
\usepackage{graphicx}
\usepackage{framed}
\usepackage[normalem]{ulem}
\usepackage{psfrag}
\usepackage{amsmath}
\usepackage{amsthm}
\usepackage{subfigure,hyperref}
\usepackage{amssymb}
\usepackage{amsfonts}
\usepackage{enumerate}
\usepackage[utf8]{inputenc}
\usepackage[top=1.2 in,bottom=1.1in, left=1.2 in, right=1.2 in]{geometry}
\usepackage{algorithm,algorithmic}
\usepackage{framed} 
\usepackage{mdframed}
\usepackage{tcolorbox}
\usepackage{mathtools}
\usepackage{url}
\usepackage{booktabs}
\usepackage{amssymb,bm,mathrsfs}
\usepackage{authblk}
\usepackage[misc]{ifsym}
\allowdisplaybreaks
\numberwithin{equation}{section}

\theoremstyle{definition}

\title{Wasserstein-Wasserstein Auto-Encoders}

\author[1]{Shunkang Zhang}
\author[2]{Yuan Gao}
\author[3]{Yuling Jiao\thanks{
Yuling Jiao (yulingjiaomath@whu.edu.cn)}}
\author[4]{Jin Liu}
\author[1]{Yang Wang}
\author[1]{Can Yang\thanks{
Can Yang (macyang@ust.hk)}}

\affil[1]{School of Mathematics and Statistics, Xi’an Jiaotong University}
\affil[2]{School of Statistics and Mathematics, Zhongnan University of Economics and Law}
\affil[3]{Department of Mathematics, The Hong Kong University of Science and Technology}
\affil[4]{Center of Quantitative Medicine Duke-NUS Medical School}

\date{Feb 24, 2019}

\begin{document}

\maketitle

\begin{abstract}
To address the challenges in learning deep generative
models (e.g.,the blurriness of variational
auto-encoder and the instability of training generative
adversarial networks, we propose a novel deep
generative model, named Wasserstein-Wasserstein
auto-encoders (WWAE). We formulate WWAE as
minimization of the penalized optimal transport between
the target distribution and the generated distribution.
By noticing that both the prior $P_Z$ and
the aggregated posterior $Q_Z$ of the latent code Z
can be well captured by Gaussians, the proposed
WWAE utilizes the closed-form of the squared
Wasserstein-2 distance for two Gaussians in the optimization
process. As a result, WWAE does not
suffer from the sampling burden and it is computationally
efficient by leveraging the reparameterization
trick. Numerical results evaluated on multiple
benchmark datasets including MNIST, fashion-
MNIST and CelebA show that WWAE learns better
latent structures than VAEs and generates samples
of better visual quality and higher FID scores than
VAEs and GANs.
\end{abstract}

\section{Introduction}
\subsection{Literature review}
\emph{Deep generative models} (DGM) have proved powerful for extracting high-level representations from real-world data such as images, audios and texts. Benefiting from probabilistic formulations and neural network architectures, DGMs allow fast sampling and play a central role in many applications, such as text to image synthesis \cite{reed16}, style transfer \cite{zhu17}, speech enhancement \cite{pascual17}, text to speech synthesis \cite{oord16}. Modern DGMs focus on mapping latent variables to fake data whose distribution is expected to closely match the real data distribution. \emph{Generative adversarial networks} (GAN) \cite{goodfellow14} and \emph{variational auto-encoders} (VAE) \cite{kingma13} are representative work in  this category.

GANs build a two player game where the generator consecutively produces fake data to deceive the discriminator while the discriminator simultaneously improves its judgment, theoretically yielding  a min-max problem. The objective of vanilla GANs, formed as a zero-sum game, amounts to minimization of the \emph{Jensen-Shannon} (JS) divergence between the fake data distribution and the real data distributions. Recent advancements in development of GANs suggest three perspectives: (1) density ratio estimation \cite{nowozin16,uehara16}; (2) kernel two-sample tests \cite{li15,dziugaite15,sutherland16,li17,binkowski18} and (3) optimal transport \cite{arjovsky17,gulrajani17,miyato18}. Density ratio estimation relies on evaluating a certain function of density ratio, generally not well-defined for distributions whose support are low-dimensional manifolds. Kernel two-sample tests and optimal transport commonly borrow strength from the classical \emph{integral probability metrics} (IPM) \cite{muller97,srip12}, giving birth to efficient and effective methods such as  MMD GAN \cite{li15},
\cite{dziugaite15} and Wasserstein GAN \cite{arjovsky17}.

VAEs constrain the latent space with a simple prior and perform approximate maximum likelihood estimation via maximizing the corresponding evidence lower bound (ELBO). ELBO-based deep generative learning enjoys optimization stability but was disputed for generating blurry image samples.
In fact, ELBO can be decomposed into a data space fitting term and a latent space regularization term, motivating a better design of the objective function by refining either the data fitting term or the regularization term.  For example,
 Adversarial auto-encoders (AAE) \cite{makhzani15} use GANs to better regularize the aggregated posterior of latent codes. Wasserstein auto-encoders (WAE) \cite{tolstikhin18}, from the viewpoint of optimal transport, generalize AAEs with \emph{penalized optimal transport} (POT) objectives \cite{bousquet17}. Similar ideas are found in some works on disentangled representations of natural images \cite{higgins16}, \cite{kumar18}.

In this work, we propose a novel generative model named \emph{Wasserstein-Wasserstein auto-encoders} (WWAE), where the squared Wasserstein-2 distance is employed to match the aggregated posterior of latent codes with a Gaussian prior while the generative model is learned via minimizing a penalized optimal optimal transport.

\subsection{Contributions}
Our main contributions are as follows:
\begin{itemize}

\item We enrich the POT generative modeling framework by considering  minimization of Wasserstein distances that are characterized for metrizing weak convergence of probability measures \cite{weed17} in  latent code space. WWAE naturally inherits the theoretical no-blurriness generation advantage of POT-based AEs \cite{bousquet17}.

\item Instead of  using  a superfluous GAN involved in AAE or WAE-GAN \cite{tolstikhin18}, the proposed WWAE utilizes the closed-form Fr{\'e}chet distance for two Gaussians \cite{dowson82}. As a result, WWAE does not suffer from the the sampling burden in WAE-MMD \cite{tolstikhin18} and it is computationally efficient by leveraging the reparameterization trick in VAE.

\item Empirically, WWAEs generate samples of higher visual quality than VAEs and WAEs when evaluated on multiple benchmark datasets, including MNIST, fashion-MNIST and CelebA. For FID scores \cite{heusel17}, WWAE matches or outperforms VAEs and WAEs.
\end{itemize}

\section{WWAE}

\subsection{Background, Notation}\label{sec2}
Let $\{\mathbf{X}_i\}_{i=1}^N \in \mathcal{X} \subset \mathcal{R}^d$  be  independent and  identically distributed samples from an unknown target  distribution
$P_X $ living in  the probability space  $\mathcal{P}(\mathcal{X})$. We aim to learn  a  deep neural network   $G_{\theta}$ with parameter  $\theta$
  that transforms low dimensional random   Gaussian  samples  $Z \in  \mathcal{Z} \subset \mathcal{R}^{\ell}$
 into samples  from $P_X$. Let $P_G$  denote the distribution of   $G_{\theta} (Z)$
 and the  Wasserstein distance $\textrm{W}_c(P_{X}||P_{G})$ (optimal transport loss) \cite{villani2008optimal} to measure the discrepancy between $P_X$ and $P_G$.
Recall the Kantorovich's formulation of the  Wasserstein distance \cite{villani2008optimal}:
\begin{equation*}
\textrm{W}_c(P_{X}||P_{G}) = \inf_{\gamma\in \mathcal{C}(X\sim P_X,Y\sim P_G)}\{\mathbf{E}_{(X,Y)\sim \gamma }[\|X-Y\|^{c}_{c}]\},
\end{equation*}
where $\mathcal{C}(X\sim P_X,Y\sim P_G)$ is the coupling  set of
all joint distributions of $(X,Y)$ with marginals  $P_X$ and $P_G$ respectively, and  $\|X-Y\|_{c}$ with $c\geq 1$ denotes the $c$ norm on $\mathcal{R}^d$.
 Under the mild condition \cite{bousquet17}, by parameterizing the coupling set, the Wasserstein distance $\textrm{W}_c(P_{X}||P_{G})$  can be equivalently reformulated as follows:
\begin{equation}\label{wde}
\textrm{W}_c(P_{X}||P_{G}) = \inf_{Q: Q_Z = P_Z} \mathbf{E}_{P_X} \mathbf{E}_{Q(Z|X)}[\|X-G_{\theta}(Z)\|_{c}^{c}],
\end{equation}
where $P_Z$ denote the Gaussian distribution in the latent space $\mathcal{Z}$, and $Q_Z$ is the aggregated posterior  distribution of $Z$ when $X\sim P_X$, $Z\sim Q(Z|X)$, i.e., $q_Z (z) = \int p_X(x) q(z|x) dx $ with $q_Z (z)$, $p_X(x)$, and $q(z|x)$ being the corresponding densities of $Q_Z$, $P_X$, and $Q(Z|X)$, respectively.
The POT loss $\widetilde{\textrm{W}}_{c}(P_{X}||P_{G})$ is the Lagrangian version of \eqref{wde} to handle the constraint $Q_Z = P_Z$, i.e.,
\begin{align}
&\widetilde{\textrm{W}}_{c}(P_{X}||P_{G}) \\ \nonumber
= &\inf_{Q} \mathbf{E}_{P_X} \mathbf{E}_{Q(Z|X)} [\|X-G_{\theta}(Z)\|_{c}^{c}] + \lambda\cdot \mathcal{D}(Q_Z || P_Z),\label{pwde}
\end{align}
where $\mathcal{D}(Q_Z || P_Z) $ is a metric on  probability space $\mathcal{P}(\mathcal{Z}),$ and $\lambda > 0$ is the regularization  parameter.

\subsection{WWAE model and algorithm}
In this section, we propose a new deep generative model named \emph{Wasserstein-Wasserstein auto-encoders} (WWAE) in the framework of POT, where  $\mathcal{D}(Q_Z || P_Z)$ is chosen as the  Wasserstein distance $\textrm{W}_2(Q_Z||P_Z)$.  The reasons that we propose to use Wasserstein distance $\textrm{W}_2(Q_Z||P_Z)$ as a regularizer are as follows:
\begin{itemize}
\item The Wasserstein distance is a weak metric \cite{weed17} that gives more regularity of the objective functions, yielding a more stable algorithm for training \cite{arjovsky17}.
\item In deep generative models, the prior $P_Z$ is often chosen as Gaussian. We also notice that the aggregated posterior $Q_Z$ can also be approximated as Gaussian. In the mini-batch training procedure, we first sample $n$ samples  $ X_i, i= 1,..., n$ from $P_X$. By using the reparametric trick, the corresponding latent codes $\tilde{Z}_{i}$ sampled from $Q(Z|X_i)$ are Gaussian.
     Hence,
    $q_Z(z)=\int p_X(x) q(z|x) dx \approx \sum^n_{i=1} q(z|X_i)/n \approx \sum^n_{i=1} \tilde{Z}_i /n  $ corresponds to a summation of a small number of encoded latent Gaussian codes which are approximately independent. The Wasserstein distance between two Gaussians is the so called the Fr{\'e}chet distance  which has a closed-form representation  using their means and covariance matrices  \cite{dowson82}  and can be estimated  efficiently from samples.
\end{itemize}
Here we recall
that
\begin{align*}
\textrm{W}_2(P ||Q) &= \|\mu_{P}-\mu_{Q}\|_2^2+ \textrm{Trace}({\Sigma}_{P})+ \textrm{Trace}({\Sigma}_{Q}) \\
&-2 \textrm{Trace}({\Sigma}^{\frac{1}{2}}_{P}{\Sigma}^{\frac{1}{2}}_{Q}),
\end{align*}
where $P$ and $Q$ are Gaussian with means and covariance matrices  $(\mu_P,{\Sigma}_{P})$ and $(\mu_Q,{\Sigma}_{Q})$, respectively.

Based on the above considerations, we propose to optimize the following objective function to learn WWAE:
\begin{equation}\label{wwad}
\min_{G_{\theta},Q_{\phi}} \mathbf{E}_{P_X} \mathbf{E}_{Q_{\phi}(Z|X)} [\|X-G_{\theta}(Z)\|_{2}^{2}] + \lambda \cdot \textrm{W}_2(Q_Z||P_Z),
\end{equation}
where we parameterize the  encoded distribution $Q(Z|X)$ using another deep neural network $Q_{\phi}$ with parameter $\phi$. We propose the following algorithm to train
WWAE model \eqref{wwad}.
\begin{itemize}
\item Specify regularization parameter $\lambda$ and batch-size $n$. Initialize the parameters of the encoder $Q_{\phi}$
decoder and generator $G_{\theta}$.
\item Loop
\begin{itemize}
\item Sample mini-batch  ${X_i}$ from $P_X$ and  sample ${Z_i}$ from $P_Z$, ${i = 1,...,n}$. Compute sample mean  $\hat{\mu}_{Z}$ and  sample covariance matrix $\hat{\Sigma}_{Z}$ using ${Z_i}s$.
\item Sample mini-batch ${\tilde{Z}_{i}},$ from $Q_{\phi}(Z|X_i)$, ${i = 1,...,n}$. Compute sample mean  $\hat{\mu}_{\tilde{Z}}$ and  sample covariance matrix $\hat{\Sigma}_{\tilde{Z}}$ using ${\tilde{Z}_i}s$.
\item Update $\phi,\theta$ via descending
\begin{align*}
&\frac{1}{n} \sum_{i=1}^{n}\|X_i - G_{\theta}(\tilde{Z}_{i})\|_2^{2} + \lambda [\|\hat{\mu}_{Z}-\hat{\mu}_{\tilde{Z}}\|_{2}^{2} \\
&+ \textrm{Trace}(\hat{\Sigma}_{Z})+ \textrm{Trace}(\hat{\Sigma}_{\tilde{Z}}) - 2 \textrm{Trace}(\hat{\Sigma}^{\frac{1}{2}}_{Z}\hat{\Sigma}^{\frac{1}{2}}_{\tilde{Z}})]
\end{align*}
with Adam \cite{adam14}.
\end{itemize}
\item End Loop
\end{itemize}
By using the reparametric trick the latent codes ${\tilde{Z}_{i}}$  and the estimated $\hat{\mu}_{\tilde{Z}}$ and $\hat{\Sigma}_{\tilde{Z}}$ are all functions of the neural network parameter $\phi$.
Therefore, the minimization problem in the last step in our above algorithm can be done via calling the stochastic gradient gradient solver such as   the Adam \cite{adam14}.

\subsection{Related works and discussion}
The blurriness of VAEs is caused by  the combination of  the Gaussian decoder   and the regularization term in VAEs  see Section 4.1 in  \cite{bousquet17} for detail argument.
 The Gaussian decoder is induced by the  reparametric trick, which can not be avoided. The regularization term in VAEs measures  the discrepancy between the   marginal encoded distribution and the prior  distribution.
 To reduce the
blurry of VAEs, much attention has been paid to find a better
regularization term.

Along this line, some related works have been proposed in
the framework of POT, aiming to improve the performance of
deep generative models via refining the regularizer.
 Adversarial
auto-encoders (AAE)  \cite{makhzani15} utilize
GANs loss to regularize the aggregated posterior of latent
codes. Wasserstein auto-encoders (WAE) \cite{tolstikhin18}, reformulated the AAEs into POT objectives \cite{bousquet17}, giving more insightful understanding of AAEs.
 \cite{tolstikhin18}, proposed using the maximum mean
discrepancy (MMD) \cite{mmd07} between the
aggregated posterior distribution and the prior latent distribution
as the regularizer in (2). We enrich the POT generative
modeling framework by considering the Wasserstein distances
as a regularizer in the latent code space. The advantage
of the proposed WWAE over superfluous GAN / MMD based
regularizer in AAE or WAE-GAN \cite{tolstikhin18} /
WAE-MMD \cite{tolstikhin18} is that the closed-form
representation makes the computation more efficient and thus
overcomes the sampling burden in WAE-MMD.
 
\section{Experiment}
\label{others}

In this section, we performed experiments to evaluate the proposed WWAE and compared its experimental results with some other closely related deep generative models. The experimental settings are presented in Section \ref{setting}. The visualization-based quality illustration and the numerical quality analysis are shown in Section \ref{quality} and Section \ref{numerical}, respectively.

\subsection{Experimental Settings}
\label{setting}
\textbf{Dataset:} We trained WWAE for image generation and reconstruction on the MNIST \cite{mnist}, Fashion-MNIST \cite{fashion_mnist}and CelebA \cite{celebA}  datasets, where the size of training instances were 30K, 30K,  200K. For the CelebA dataset, we cropped and resized the image into $64 \times 64$ as the previous research. 

\begin{figure}[H]
\label{figure1}
  \centering
  \subfigure[Real image]{\includegraphics[width=5cm, height=5cm]{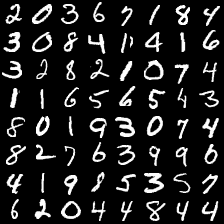}}
  \subfigure[Real image]{\includegraphics[width=5cm, height=5cm]{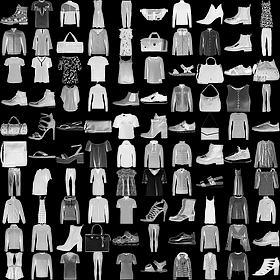}}
    \subfigure[Reconstructed image]{\includegraphics[width=5cm, height=5cm]{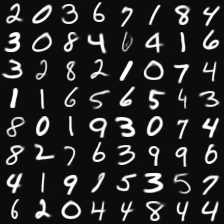}}
  \subfigure[Reconstructed image]{\includegraphics[width=5cm, height=5cm]{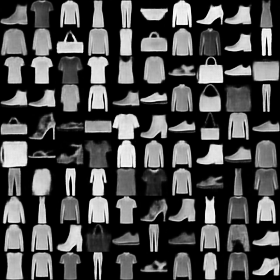}}
    \subfigure[Generated image]{\includegraphics[width=5cm, height=5cm]{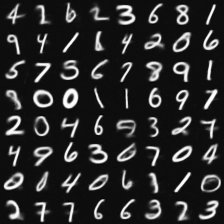}}
 \subfigure[Generated image]{\includegraphics[width=5cm, height=5cm]{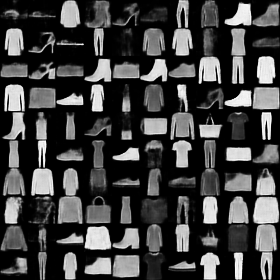}}
  \caption{\label{figure1} Real samples, reconstructed samples and generated samples from WWAE with batch size 64 on the MNIST and Fashion-MNIST datasets.}
\end{figure}

\noindent \textbf{Network architecture:} In our experiments, we adopted the architecture of traditional convolution neural networks used in VAE to design decoder $G_{\theta}$ and encoder $Q_{\phi}$. After a few experiments, however, we found that the traditional VAE architecture was too shallow to capture the high level features in images, as well as suffered from a slower rate of convergence. Therefore, we added several residual blocks \cite{residual} in both the encoder and the decoder to extract more high-level features and improve the image quality. Specifically, we added two residual blocks between every two convolution layers in our encoder and decoder while maintained other network structure like DCGAN.
\begin{figure}
\label{figure2}
  \centering
  \subfigure[Manifold (WWAE)]{\includegraphics[width=5cm, height=5cm]{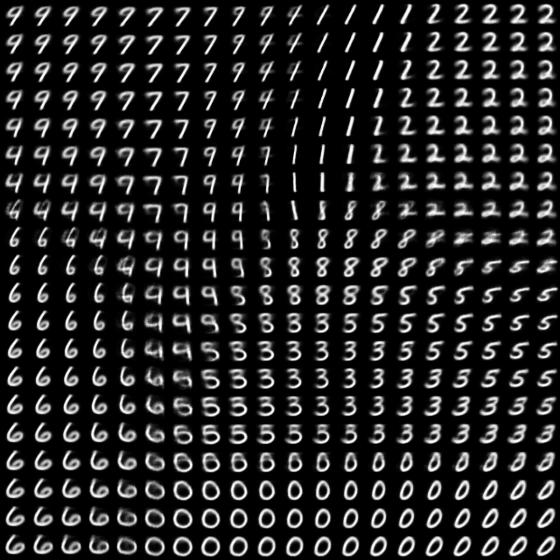}}
  \subfigure[Label plot (WWAE)]{\includegraphics[width=5cm, height=5cm]{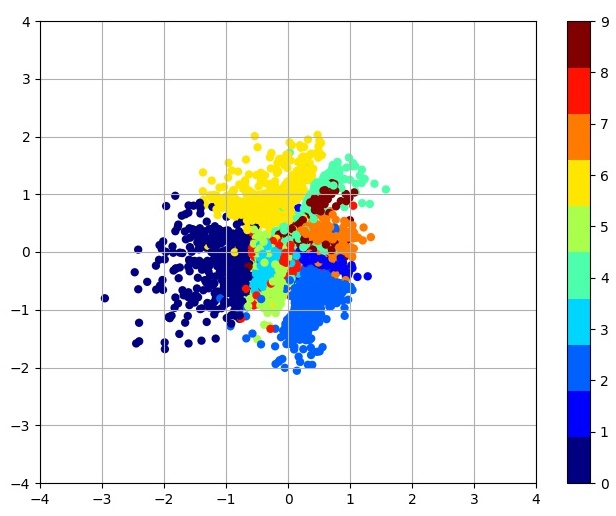}} \\
   \subfigure[Manifold (VAE)]{\includegraphics[width=5cm, height=5cm]{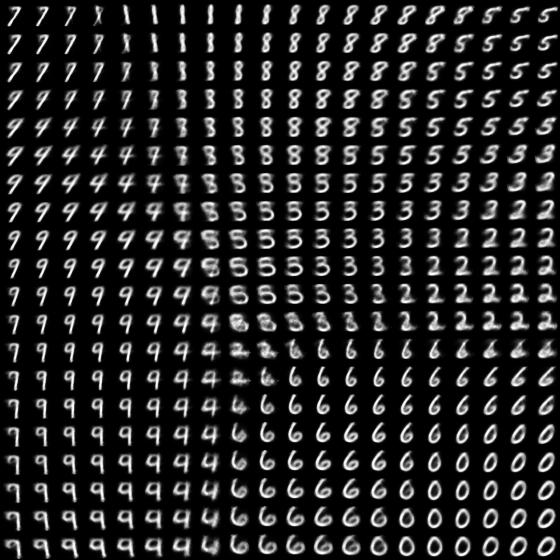}}
  \subfigure[Label plot (VAE)]{\includegraphics[width=5cm, height=5cm]{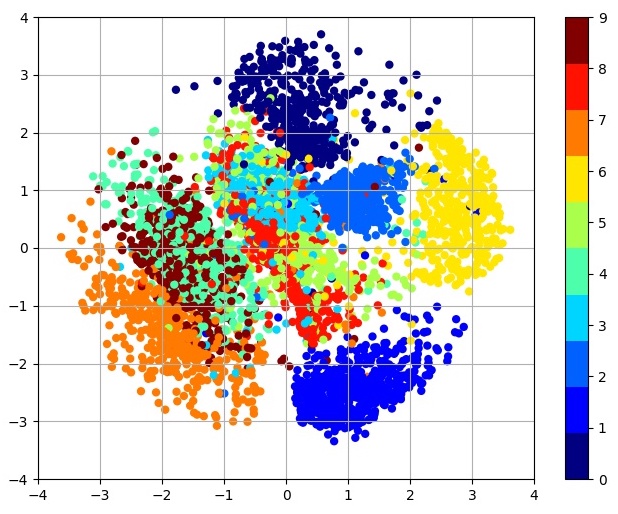}}
  \caption{\label{figure2} Manifolds learned by WWAE (a) and VAE (c) on MNIST, and distributions of latent codes learned by WWAE (b) and VAE (d) with labels.}
  \end{figure}
\noindent \textbf{Hyper-parameters setting:} We used the  Adam optimizer \cite{adam14} with starting learning rate $r=0.005$, $\beta_{1} =0.5, \beta_{2} =0.9$. We decayed our learning rate every 10K by 0.9 on MNIST, Fashion-MNIST and every 20K by 0.9 on CelebA dataset, respectively.  We used  batch normalization layers after each convolution layer except the final one, and  used Relu activation function in our neural network. We choose batch size as 64 on MNIST and Fashion-MNIST, while increased batch size to 100 on CelebA.

\subsection{Qualitative Analysis}
\label{quality}

Due to the relatively simple structure of images in MNIST and Fashion-MNIST, it is easier for WWAE to  achieve  a good performance on both reconstructed images and generated images. As shown in Figure~\ref{figure1}, images generated by WWAE not only have clear details but also achieved a fairly good diversity.

\begin{figure}
\label{figure3}
  \centering
  \subfigure[Real image]{\includegraphics[width=8cm, height=6.5cm]{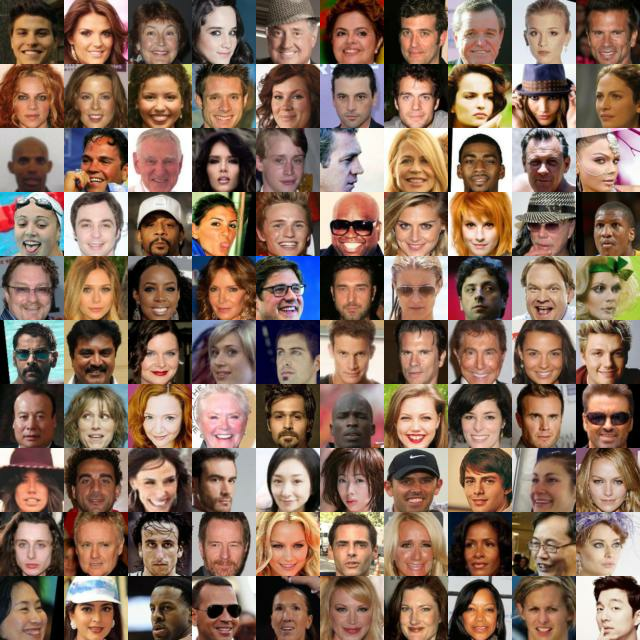}}
  \subfigure[Reconstructed image]{\includegraphics[width=8cm, height=6.5cm]{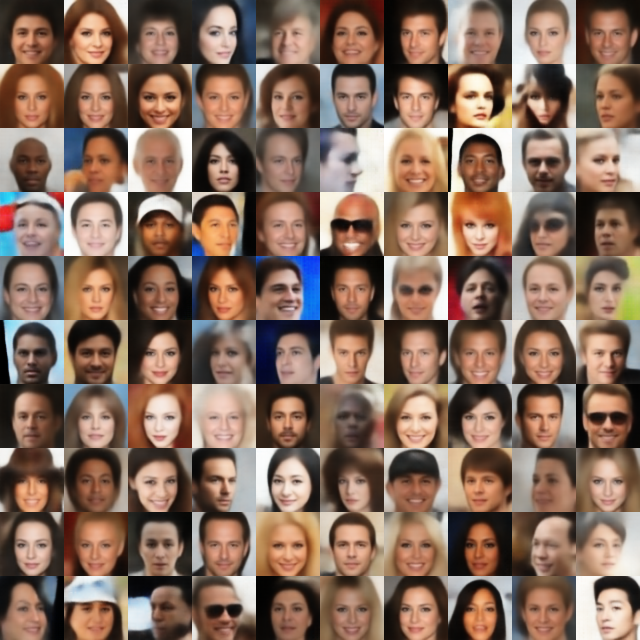}}
  \subfigure[Generated image]{\includegraphics[width=8cm, height=6.5cm]{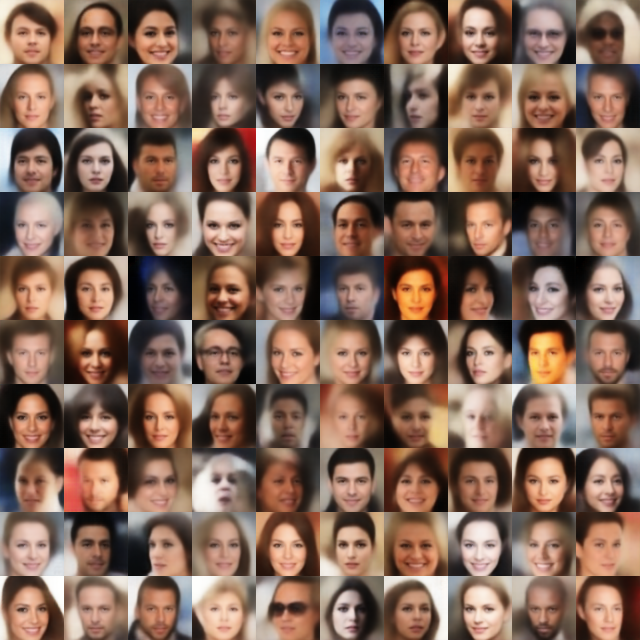}}
    \caption{\label{figure3}Real samples, reconstructed samples and generated samples from WWAE with batch size 100 on the CelebA dataset.}
\end{figure}

\begin{figure}
\label{figure4}
  \centering
  \includegraphics[width=14cm, height=9cm]{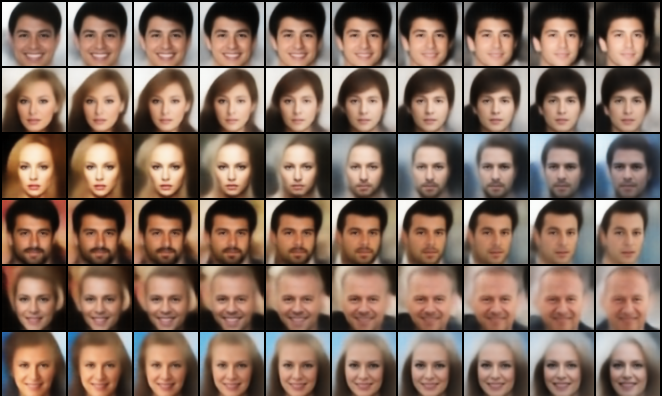}
  \caption{\label{figure4}Latent space interpolation learned by WWAE.}
\end{figure}

Figure~\ref{figure2} shows the learned manifolds based on the MNIST dataset when dimension of latent space equals to 2. Both VAE and our proposed WWAE can learn a manifold to represent the data structure. For the manifold learned by VAE, Figure \ref{figure2} (d) shows that there are large holes between two different labels (e.g., digits 0 and 6). If an image whose latent code is collected from a hole, then the image can be blurry because its latent code would have to be averaged from its neighbors (e.g., the average of the latent codes of 0 and 6).  This artifact has been greatly reduced for the manifold learned by WWAE shown in Figure 2 (b). The better performance of WWAE could be attributed to Wasserstein 2 distance as the regularizer. As a result, the generated image from WWAE can be less blurry than that from VAE. We also evaluated our model on a more complex RGB dataset CelebA. The results are shown in Figure~\ref{figure3}. Furthermore, the latent space of WWAE on the CelebA dataset is visualized in Figure \ref{figure4}. We can see a smooth transform between two different faces, indicating a continuous and well-structured latent space representation has been learned by WWAE.

\par \textbf{Model comparison:} In order to demonstrate the  sampling  quality and  convergence, we first compared our WWAE model with DCGAN which was one of most popular deep generative models in various applications and researches. Regarding the encoder-decoder framework, we made comparison between our WWAE model and VAE, WAE-MMD. In  Figure~\ref{figure5}, we show that WWAE can generative more realistic images without too much blur and distortion. However, the generated images from VAE are often blurry while the images generated from DCGAN and WAE-MMD are not quite stable, involving distortions and artifacts.
We also observed the instability issue during the DCGAN training process, i.e., DCGAN often collapses if the training time is relatively long.
%

\subsection{Quantitative Analysis}
\label{numerical}
\par To quantitatively measure the quality and diversity of generated samples, we compute the Fr{\'e}chet Inception Distance \cite{fid} score on CelebA dataset. The FID score is a numerical indicator which is more robust to noise than inception score and more sensitive to model collapse. It measures the quality of generated image samples by comparing the statistics of generated samples to real samples.

\par  We first randomly selected 10k real images from the real datasets, serving  as the standard to compute the FID score. 
 As shown in Table \ref{table1}, we see that the proposed WWAE can achieve better FID score on all of the three datasets than  WAE-MMD and VAE. Although DCGAN slightly outperformed WWAE on MNIST and Fashion MNIST, it failed to be stably applied to some more  complex dataset, such as CelebA. To show a detailed comparison among WWAE, WAE-MMD, VAE and DCGAN, we calculated their FID scores every 5K iterations based on their generated 10K images. As shown in Figure \ref{figure6}, WWAE and WAE-MMD perform better than VAE, while DCGAN is confirmed to be unstable during the training process.

\makeatletter\def\@captype{table}\makeatother

\begin{table}
\caption{\label{table1} FID scores on MNIST, F-MNIST, and CelebA}
  \label{table1}
\centering
 \begin{tabular}{c c c c}
    \toprule
    \cmidrule(r){1-2}
    Model     & MNIST & F-MNIST & CelebA   \\
    \midrule
    Real data & 1 & 1 & 2  \\
    WWAE &  45 & 51 & 55      \\
    WAE-MMD     & 50 & 54 & 55    \\
    VAE     & 47 & 56 & 66      \\
    DCGAN & 35 & 44 &  --- \\
    \bottomrule
  \end{tabular}
  \end{table}

\par We note that WWAE and WAE-MMD have a very similar FID score on the CelebA dataset. Noticing that FID score is only a numerical critieria measuring the quality of generated images, it may not fully represent the visual quality by human eyes. In Figure \ref{figure5}, we show the generated images from the four different methods to compare their visual quality. Clearly, the quality of images generated by WWAE is better than that of the other methods.

\begin{figure}\label{figure5}
  \centering
  \subfigure[WWAE]{\includegraphics[width=6cm, height=6cm]{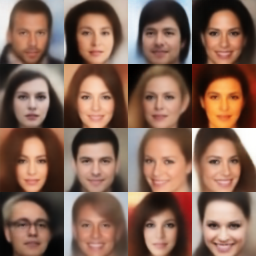}}
  \subfigure[VAE]{\includegraphics[width=6cm, height=6cm]{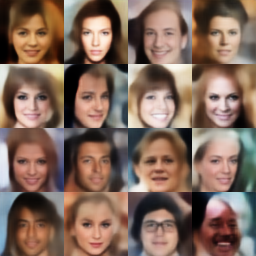}}
  \subfigure[DCGAN]{\includegraphics[width=6cm, height=6cm]{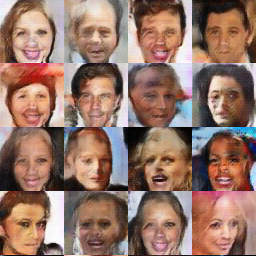}}
    \subfigure[WAE-MMD]{\includegraphics[width=6cm, height=6cm]{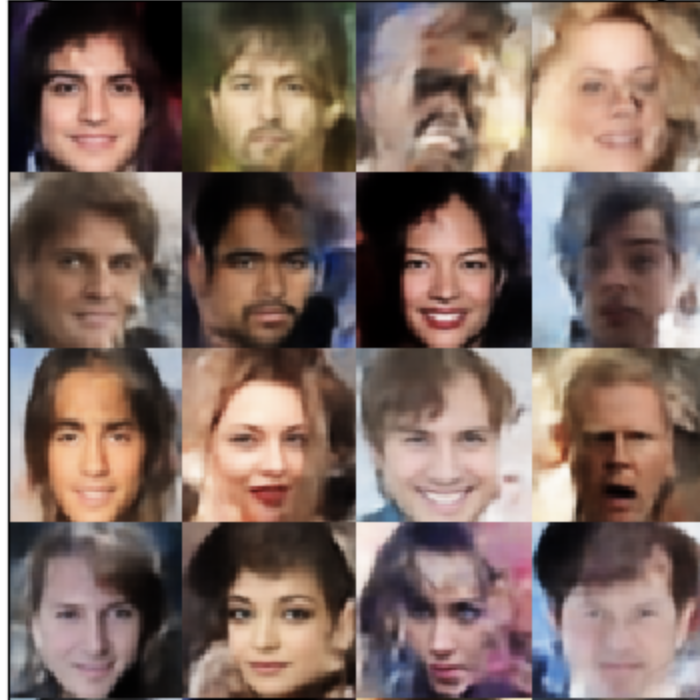}}
    \caption{\label{figure5}Comparison between generated samples from WWAE, DCGAN, VAE and WAE-MMD.}
\end{figure}

\section{Conclusion}
In this paper, we propose a novel deep generative model named \emph{Wasserstein-Wasserstein auto-encoders} (WWAE).
We   approximate the optimal transport loss of the target distribution and the generated distribution via  penalized optimal transport where
the squared Wasserstein-2 distance between the aggregated posterior of latent codes and  a Gaussian prior is served as the penalty.
 WWAE reduces  the  blurriness  of VAE and improve the stability of training deep generative models.
 Numerical results evaluated on multiple benchmark datasets including MNIST, fashion-MNIST and CelebA show that  WWAE learns better latent structures then VAEs  and generates samples of better visual quality and higher FID scores than VAEs and GANs. that  WWAE learns better latent structures then VAEs  and generates samples of better visual quality and higher FID scores than VAEs and GANs.

\makeatletter\def\@captype{figure}\makeatother

\begin{figure}\label{figure6}
\centering
\includegraphics[width=9cm, height=9cm]{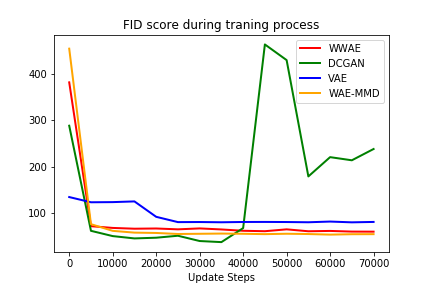}
\caption{\label{figure6} FID score comparison on celebA}
\end{figure}

\bibliographystyle{plain}
\bibliography{ref}

\end{document}